\documentclass[conference]{IEEEtran}
\IEEEoverridecommandlockouts
\usepackage{amsmath,amssymb,amsfonts}
\usepackage{balance}

\usepackage{upgreek} %upgreek used for \wordsize command
\usepackage{mathtools} % for \coloneqq
\usepackage[noend]{algorithmic} %[noend] option omits "end if" and "end for" from algorithms
\usepackage{algorithm}
\usepackage{graphicx}
\usepackage{textcomp}
\usepackage{xcolor}
\def\BibTeX{{\rm B\kern-.05em{\sc i\kern-.025em b}\kern-.08em
    T\kern-.1667em\lower.7ex\hbox{E}\kern-.125emX}}
    
\usepackage[commentmarkup=uwave,draft]{changes}
\usepackage{tablefootnote}

\usepackage{url}

\usepackage{breakurl}

\usepackage{subcaption}

\makeatletter
\let\NAT@parse\undefined
\makeatother
\usepackage[breaklinks,hidelinks,colorlinks=true,linkcolor=blue,citecolor=blue,pagebackref=true]{hyperref}

\usepackage[firstpage=true,color=black]{background}

\SetBgContents{Camera Ready Manuscript}% Set contents
\SetBgPosition{current page.east}% Select location
\SetBgVshift{0.6cm}% Add vertical shift (results in a shift in x direction due to rotation)
\SetBgOpacity{0.5}% Select opacity
\SetBgAngle{90.0}% Select rotation of logo
\SetBgScale{1.5}% Select scale factor of logo

\definechangesauthor[name={KTB}, color=red]{KTB}
\definechangesauthor[name={Evelio}, color=green]{Evelio}
\definechangesauthor[name={Gilbert}, color=orange]{Gilbert}
\definechangesauthor[name={Jose}, color=purple]{Jose}

\begin{document}

\title{Path Planning using Reinforcement Learning:\\ A Policy Iteration Approach}

%% Authors List

% \author{
% \IEEEauthorblockN{
% Saumil Shivdikar\IEEEauthorrefmark{1},
% John Doe\IEEEauthorrefmark{2},
% }
% \IEEEauthorblockA{
% \IEEEauthorrefmark{1}Somaiya Vidyavihar University,
% \IEEEauthorrefmark{2}New University, \\
% {saumil}@somaiya.edu,
% {john.doe}@newuni.edu, 
% }
% }

\author{\IEEEauthorblockN{Saumil Shivdikar}
\IEEEauthorblockA{\textit{Electronics Dept.} \\
\textit{Somaiya Vidyavihar University}\\
Mumbai, India \\
{saumil.shivdikar}@somaiya.edu}
\and
\IEEEauthorblockN{Jagannath Nirmal}
\IEEEauthorblockA{\textit{Electronics Dept.} \\
\textit{Somaiya Vidyavihar University}\\
Mumbai, India \\
{jhnirmal}@somaiya.edu}
}

\maketitle

\begin{abstract}

With the impact of real-time processing being realized in the recent past, the need for efficient implementations of reinforcement learning algorithms has been on the rise.
Albeit the numerous advantages of Bellman equations utilized in RL algorithms, they are not without the large search space of design parameters.

This research aims to shed light on the design space exploration associated with reinforcement learning parameters, specifically that of \emph{Policy Iteration}.
Given the large computational expenses of fine-tuning the parameters of reinforcement learning algorithms, we propose an auto-tuner-based ordinal regression approach to accelerate the process of exploring these parameters and, in return, accelerate convergence towards an optimal policy.
Our approach provides $1.82 \times$ peak speedup with an average of $1.48 \times$ speedup over the previous state of the art.

\end{abstract}

\begin{IEEEkeywords}
Design Space Exploration, Policy Convergence, Markov Design Process, Maze Solving
\end{IEEEkeywords}

\section{Introduction}
When problems of Markovian nature (that incorporate the Markov Design Process (MDP)) have been introduced, \emph{Policy Iteration} and \emph{Value Iteration} algorithms have been at the forefront of preferred choice as a solution~\cite{sutton1999reinforcement, baruah_gnnmark_2021}.
These algorithms fall under the domain of \emph{Dynamic Programming}, which are capable of searching for computationally optimal policies~\cite{sutton_reinforcement_2018}.

A major drawback of these approaches lies in the assumption that the surrounding environment is a perfect model~\cite{littman1995learning}.
Despite the flawed assumption of an ideal environment, Dynamic Programming remains integral to developing solutions to simulated environments~\cite{vakil2021learning, srinivasan_dynamic_2016}.

Dynamic Programming algorithms are reward-inspired solutions.
By creating a system that is rewarded for inching closer toward the objective and penalized for actions that do not benefit the user, it is possible to create a policy that finds a close to the ideal solution to reach objective functions while satisfying all criteria.
One of the challenges in creating such a system lies in setting the values of the parameters that describe the rewards, penalty, and learning rates of our problem statement.
With the area to explore being exponentially large, a brute-force approach would be far from ideal for this problem. Often such an approach yields non-optimal solutions that are prone to converge upon local minimas~\cite{shivdikar_automatic_2015}.

\begin{figure}[htbp]
	\centering
	\includegraphics[width=0.48\textwidth]{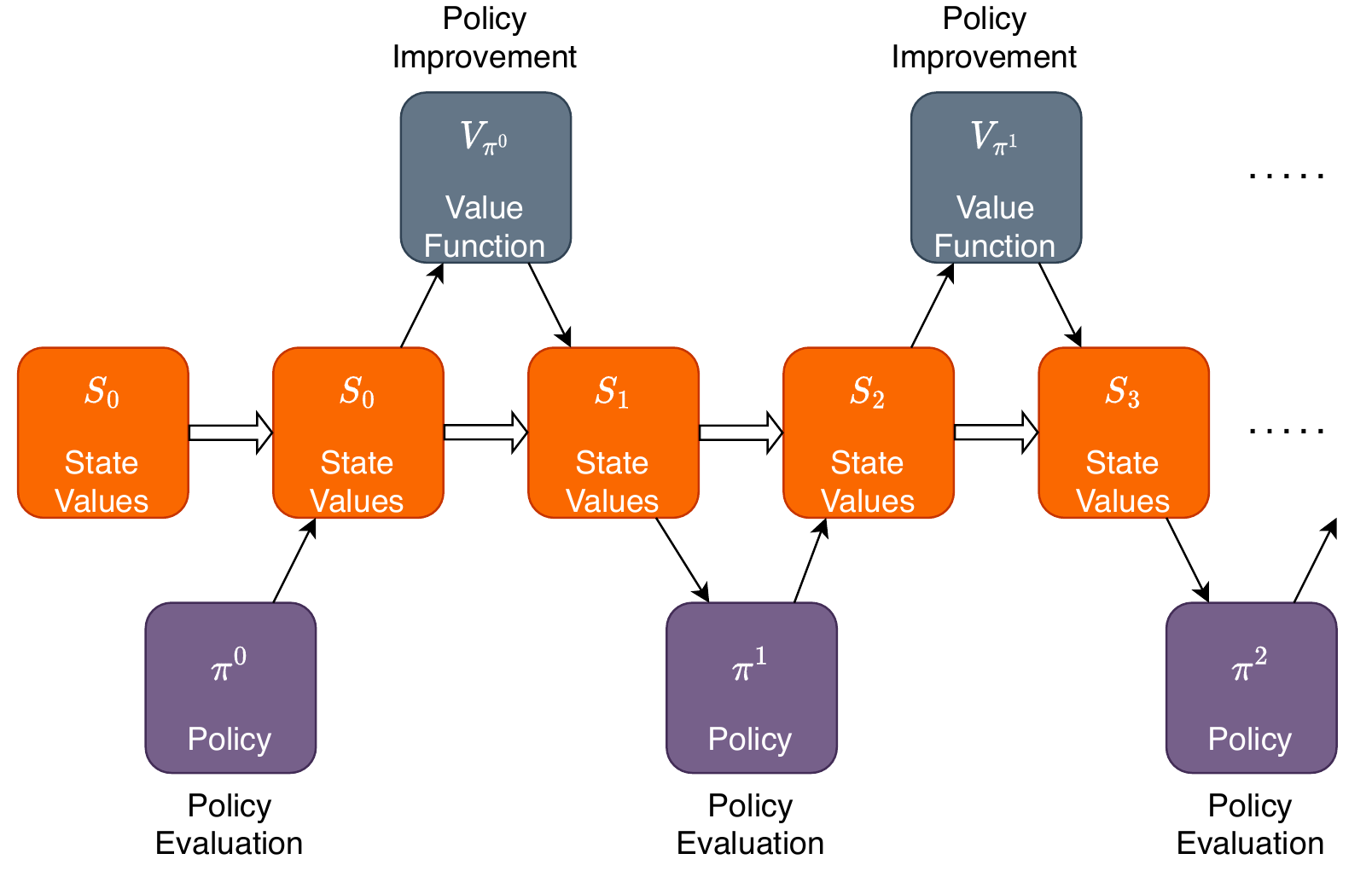}
	% \vspace{-0.7em}
	\caption{Policy Iteration Flowchart: A visual representation of the policy iteration algorithm}
	%\caption{FHE protects against network insecurities in untrusted cloud services, enabling users to securely offload sensitive data}
	\label{fig:pi_flow}
\vspace{-1.9em}
\end{figure}

This work contributes toward improving Policy Iteration in the following ways:
\begin{enumerate}
    \item We provide an exhaustive analysis of previous attempts at optimizing search space exploration
    \item We evaluate the permutations of different values of learning parameters of the Policy Iteration algorithm
    \item We propose a novel approach based on ordinal regression auto-tuning to accelerate the design-space exploration of reward parameters
    \item We evaluate our proposed solution across various problem statements and provide an average acceleration of $1.48\times$ over heuristic approaches
\end{enumerate}

The rest of this research is compartmentalized as follows;
Section~\ref{sec:policy_iter} provides a high-level explanation to understand \emph{Policy Iteration}, this is followed by Section~\ref{sec:auto_tuner} that formalizes our approach to reduce the search space of design space exploration with the use of Auto-Tuner. Section~\ref{sec:exp_setup} describes the nature of the experiments we used to evaluate our proposed solution. Section~\ref{sec:results} describes our findings which are later compared in Section~\ref{sec:related_work}. Finally, we provide a brief overview of our results in Section~\ref{sec:conclusion}.

\section{Policy Iteration}
\label{sec:policy_iter}
Policy iteration falls under the umbrella of \emph{Dynamic Programming} algorithms.
These algorithms aim to solve problems in a simulated model environment as a Markov Design Process (MDP).
Albeit Policy Iteration is a practical algorithm in exploring close to an optimal solution, it is associated with hefty computational expenses that scale quadratically with an increase in design space.
We start off this section with a description of the parameters before diving into the implementation of policy iteration for path planning.

\subsection{Parameter description}
\begin{table}[htb]
% \vspace{-0.6em}
	\centering % used for centering table
	% p{20mm}
	\begin{tabular}{l l} % centered columns (4 columns)
		\hline\hline %inserts double horizontal lines
		\textbf{Parameter} & \textbf{Description} \\ [1.0ex] % inserts table
		%heading
		\hline %\\ [0.5ex] % inserts single horizontal line
		  $S_t$ & State value at time $t$\\
            $R_t$ & Reward obtained at time $t$\\
            $\pi$ & Current policy \\
            $\pi_*$ & Optimal Policy\\
            $v_{\pi}(s)$ & Value function following \emph{current} policy $t$\\
            $v_{\pi_*}(s)$ & Value function following \emph{optimal} policy $t$\\
    	  $\gamma$ & The discount rate\\ [1ex]
		\hline \\ [1ex]  %inserts single line 
	\end{tabular}
	\caption{Description of parameters used in policy iteration path planning} % title of Table
	\label{tab:param_desc} % is used to refer this table in the text
\vspace{-1.0em}
\end{table}

\begin{figure}[tbp]
	\centering
	\includegraphics[width=0.48\textwidth]{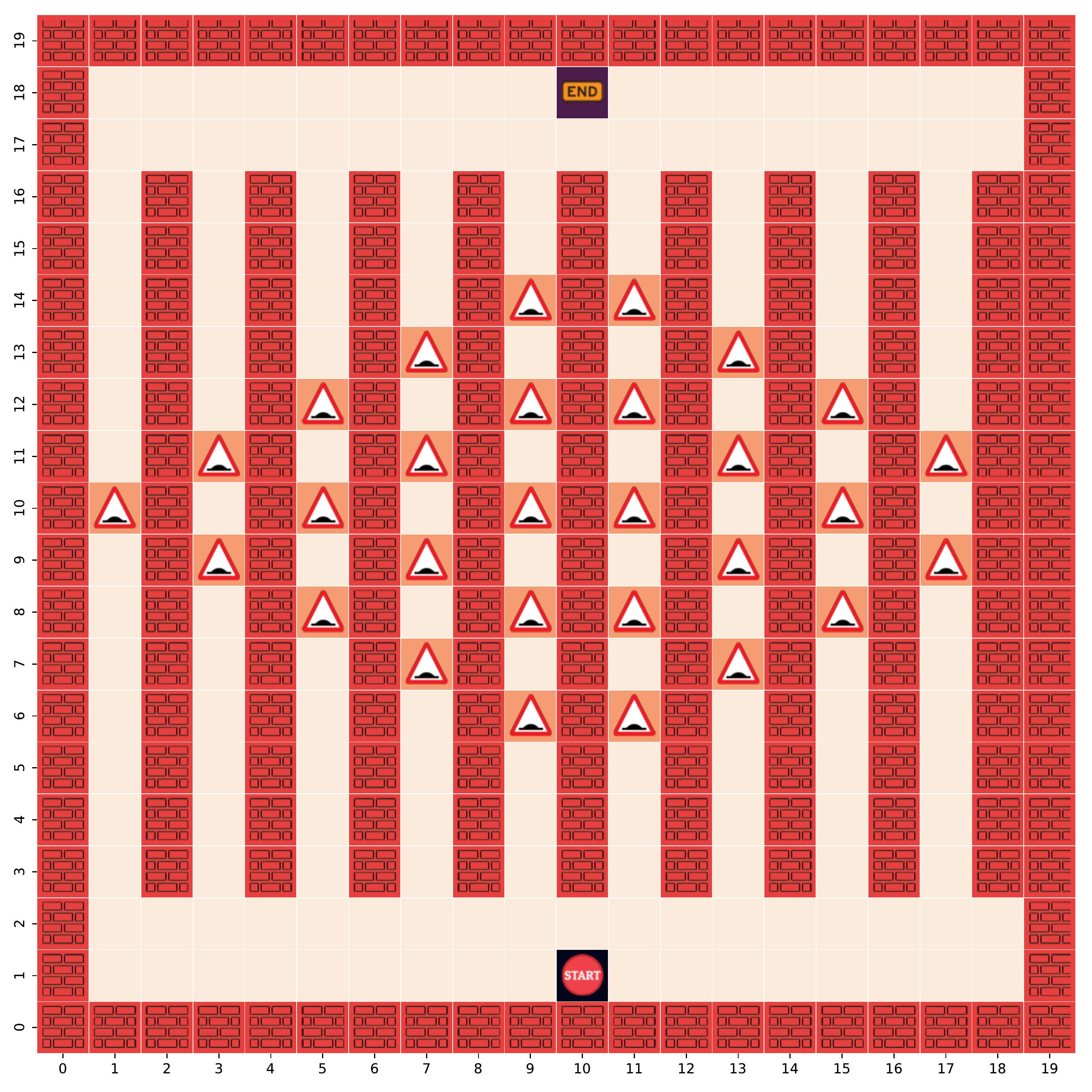}
	% \vspace{-0.7em}
	\caption{An example of a multi-lane maze with speed bumps. The challenge lies in choosing a closer lane with more speed bumps or a farther lane with fewer speed bumps.}
	%\caption{FHE protects against network insecurities in untrusted cloud services, enabling users to securely offload sensitive data}
	\label{fig:maze_empty}
\vspace{-1.9em}
\end{figure}

We describe our policy iteration approach to path planning using a toy example.
Figure~\ref{fig:maze_empty} is an example maze with multiple lanes from start to finish.
\begin{itemize}
    \item \textbf{State Values} ($S_t$): To describe our problem statement, we represent the maze with a matrix of state values (each value representing a position within the maze).
    \item \textbf{Policy} ($\pi$): A mapping of states to possible actions.
    \item \textbf{Value Function} ($v_{\pi}(s)$): The expected return obtained after following policy ($\pi$) when start location is $s$.
    \item \textbf{Reward} ($R_t$): The reward received at time $t$.
    \item \textbf{Learning Rate} ($\gamma$): The rate at which errors from the previous iteration propagate into the next iteration
\end{itemize}
We incorporate the Bellman optimality equation from~\cite{gross2016bellman} toward our path planning problem statement as seen in equations~\ref{eq:bellman_1} and~\ref{eq:bellman_2}.
\begin{equation}
\label{eq:bellman_1}
    v_*(s) = max_{a}\mathbb{E}[R_{t+1} + \gamma v_*(S_{t+1}) \ | \ S_t = s, A_t = a]
\end{equation}

\begin{equation}
\label{eq:bellman_2}
    V_{\pi}(s) = \sum_{s,r}P(s|S,\pi(s))[r + \gamma V_{\pi(s)}], s \in S
\end{equation}

for all $s \in S$, $a \in \mathcal{A}(s)$, and $s' \in \mathcal{S}^+$

\begin{figure}[tbp]
	\centering
	\includegraphics[width=0.48\textwidth]{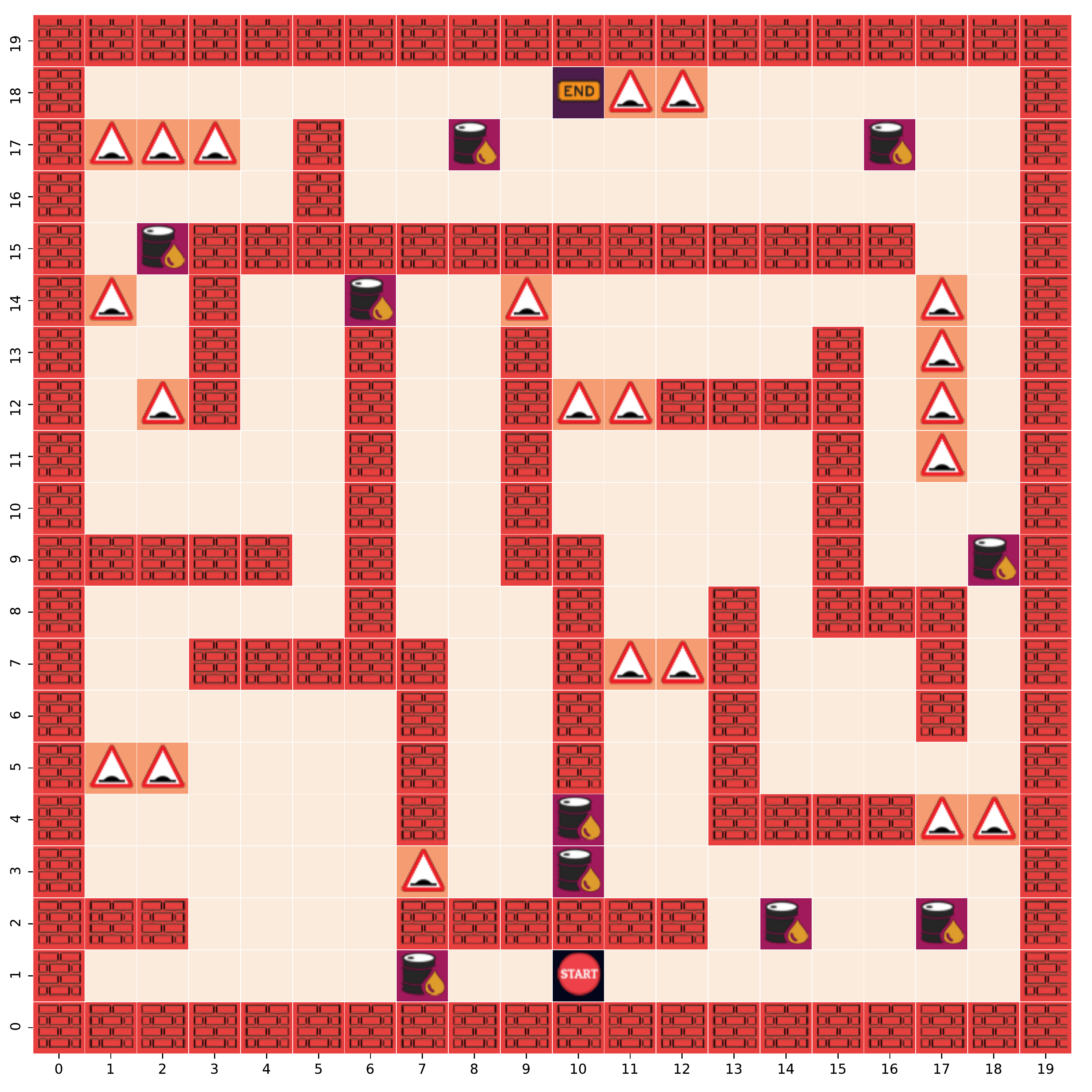}
	% \vspace{-0.7em}
	\caption{A maze with multiple types of obstacles. The challenge lies in choosing an optimal path based on the reward policy that penalizes differently for each obstacle.}
	%\caption{FHE protects against network insecurities in untrusted cloud services, enabling users to securely offload sensitive data}
	\label{fig:maze_empty_complex}
\vspace{-1.9em}
\end{figure}

Figure~\ref{fig:maze_empty_complex} is another example we incorporate to illustrate the behavior of our design space explorer under various obstacle types. This type of maze is known as a "multi-objective maze" or a "multi-modal maze" because it has multiple objectives or modes of operation. The challenge in solving such a maze is to find a path that balances the trade-offs between different types of obstacles and the associated penalties while also maximizing the overall reward. This type of problem is often solved using a combination of search algorithms and optimization techniques~\cite{jayaweera_jaxed_2021, shivdikarspeeding}.

\subsection{Policy Evaluation}

\begin{algorithm}
\caption{Policy Evaluation}\label{alg:pol_eval}
\begin{algorithmic}[1]
\REQUIRE $\pi$, the policy to be evaluated
\ENSURE $\Delta < \theta$
\WHILE{$\Delta < \theta$}{
    \STATE $\Delta \gets 0$
    \FOR{$s \in \mathbb{S}$}{
    \STATE $v \gets V(s)$
    \STATE $V(s) \gets \sum_{a}\pi(a | s)\sum_{s^{'},r}p(s^{'},r|s,a)[r+\gamma V(s^{'})]$
    \STATE $\Delta \gets max(\Delta,|v - V(s)|)$
    }\ENDFOR
}\ENDWHILE
\end{algorithmic}
\end{algorithm}

Our objective is to iteratively evaluate and improve the policy $\pi$ such that it reaches as close as possible to the optimal policy $\pi_*$.
\emph{Policy Evaluation} aims to compute the efficiency of the current policy $\pi$.
This is achieved by computing the value function $V(s)$.
While computing $V(s)$ specifically for our policy $\pi$ we refer to it as $V_{\pi}(s)$.
We start off by setting the value of $\theta$ (our threshold to halt computation)~\cite{bunn_student_2019}. 
Following this, we iteratively compute the value function $V_{\pi}(s)$ following the algorithm~\ref{alg:pol_eval}.
After reaching a $\Delta$ error value that lies within our threshold $\theta$, we stop iterating.

\emph{Policy Evaluation} is a means of verifying the efficacy of our generated policy.
It is important to point out that if our surrounding environment is statistically deterministic, we can converge to our solution directly by solving a set of $|\mathcal{S}|$ linear equations.

\subsection{Policy Improvement}

\begin{algorithm}
\caption{Policy Improvement}\label{alg:pol_improv}
\begin{algorithmic}[1]
\REQUIRE $\pi$, the policy to be improved
\ENSURE $\pi$ does not change with consecutive iterations
\STATE $ policy\mbox{--}stable \gets true$
    \FOR{$s \in \mathbb{S}$}{
    \STATE $old\mbox{--}action \gets \pi (s)$
    \STATE $\pi (s) \gets \mathrm{argmax}_a \sum_{s^{'},r}p(s^{'},r|s,a)[r+\gamma V(s^{'})]$
    \IF{$old\mbox{--}action \neq \pi (s)$}{
    \STATE $ policy\mbox{--}stable \gets false$
    }\ENDIF
    \IF{$policy\mbox{--}stable = true$}{
    \RETURN $V \approx v_* , \pi \approx \pi_*$
    }\ENDIF
    }\ENDFOR

\end{algorithmic}
\end{algorithm}

The reason for computing value function in \emph{Policy Evaluation} is to improve our policy. In this stage of \emph{Policy Improvement}, we use our computed value functions to set the parameters of our policy ($\pi_t$) to create a new policy ($\pi_{t+1}$), inching a step closer to the optimal policy ($\pi_*$).

For every single state value, we compute the value function for each action.
Based on the action that provides the maximum value function, we greedily set the policy to that particular action.
This gets repeated for every state value in the system.
Although computationally expensive~\cite{shivdikar_accelerating_2022, livesay2023accelerating}, this approach provides a greedy solution to finding the best action given the current state values.
And as with all greedy solutions, such an approach is susceptible to converging toward a local maxima while ignoring the global one.
Eager readers are directed toward algorithm~\ref{alg:pol_improv} to understand \emph{Policy Improvement}.

\subsection{Policy Iteration}

\begin{algorithm}
\caption{Policy Iteration}\label{alg:pol_iter}
\begin{algorithmic}[1]
\REQUIRE $\pi \approx \pi_*$, the optimal policy
\ENSURE $V(s) \in \mathbb{R}, \pi(s) \in \mathcal{A}(s)$ for all $s \in \mathcal{S}$
\STATE $ policy\mbox{--}stable \gets false$
    \WHILE{$policy\mbox{--}stable \neq true $}{
    \STATE Compute \textbf{Policy Evaluation}
    \STATE Compute \textbf{Policy Improvement}
    \IF{$policy\mbox{--}stable$}{
    \RETURN $V \approx v_* , \pi \approx \pi_*$
    }\ENDIF
    }\ENDWHILE
\end{algorithmic}
\end{algorithm}

% Coalescing all steps together, we turn our attention towards \emph{Policy Iteration}.
% Policy Iteration, as the name suggests, is an iterative approach to alternatingly evaluate and improve a policy. 
% This approach follows the axiom that each policy is guaranteed to be a strict improvement over the previous one (unless it is the optimal one).

Coalescing all steps together, we turn our attention towards \emph{Policy Iteration}.
Policy iteration is a common method used for solving robotic maze navigation tasks. The method involves iteratively updating the policy, or the mapping from states to actions, to improve the agent's performance. The flow of this approach can be visualized in Figure~\ref{fig:pi_flow}. In the context of robot maze solving, the policy iteration process typically begins with an initial, randomly generated policy. The agent then follows this policy while navigating the maze, and the cumulative reward is recorded. The policy is then updated based on the agent's performance, and the process is repeated. Through this iterative process, the policy gradually improves, leading to better and better performance in the maze.

This study has shown that such an approach can be highly effective in solving robotic maze navigation tasks. For example, if a robot was trained to navigate through a maze using policy iteration~\cite{dang2010efficient, intrator_missing_2017}, it could successfully find the goal in a high percentage of trials and with a high degree of efficiency compared to other algorithms. In addition to this, this approach has been used to train robots to navigate through more complex mazes with multiple goals and obstacles and has shown promising results~\cite{zhu2021deep}.

\section{Auto-tuner}
\label{sec:auto_tuner}

We employ an auto-tuner using ordinal regression for design space exploration of reward function.
Our approach of incorporating auto-tuning reinforcement learning reward parameters using ordinal regression with partial ranking is a method for adjusting the weights of the reward function in a reinforcement learning algorithm~\cite{olschanowsky2014study}. The goal is to find the optimal parameters that maximize the performance of the learning agent~\cite{kronawitter2014optimization}. The method uses ordinal regression, which is a type of supervised learning where the output is a set of ordinal categories, rather than a continuous value. The partial ranking aspect involves only comparing a subset of the actions, rather than the entire set, to determine the optimal parameters~\cite{cosenza2017autotuning, baruah_gnnmark_2021, ansel2014opentuner}. This can help to reduce computational costs and improve the efficiency of the learning process. The method is typically used in complex or high-dimensional problems where traditional methods may not be effective. This work is inspired by the contributions of Cosenza et al. in their artwork~\cite{cosenza2017autotuning}.

Assuming that $P_1$, $P_2$ $\dots$ $P_n$ are all partial rankings available in the training set~\cite{hajek2014minimax}, we can formally define the ordinal ranking, incorporating partial equations, as follows:

\begin{equation}
    \min_{{\mathrm{\mathbf{w}}},\xi \geq 0} \frac{1}{2} \mathrm{\mathbf{w}}^{T} \mathrm{\mathbf{w}} + \frac{C}{m'} \sum_{i} \sum_{(j,k) \in P_{i}} \xi_{j,k}
\end{equation}

which is subject to
\begin{equation}
\begin{aligned}
    \forall (j,k) \in P_{1} : (\mathrm{\mathbf{w}}^{T}q^{1},\mathrm{\mathbf{t_{j}}}) \geq (\mathrm{\mathbf{w}}^{T}q^{1},\mathrm{\mathbf{t_{k}}}) + 1 - \xi_{j,k} \\
    \forall (j,k) \in P_{2} : (\mathrm{\mathbf{w}}^{T}q^{2},\mathrm{\mathbf{t_{j}}}) \geq (\mathrm{\mathbf{w}}^{T}q^{2},\mathrm{\mathbf{t_{k}}}) + 1 - \xi_{j,k} \\
    ... \\
    \forall (j,k) \in P_{n} : (\mathrm{\mathbf{w}}^{T}q^{n},\mathrm{\mathbf{t_{j}}}) \geq (\mathrm{\mathbf{w}}^{T}q^{n},\mathrm{\mathbf{t_{k}}}) + 1 - \xi_{j,k}
\end{aligned}
\end{equation}
where $m' = |\bigcup_{i}P_{i}|$ and $n = |Q|$

\begin{figure}[tbp]
	\centering
	\includegraphics[width=0.48\textwidth]{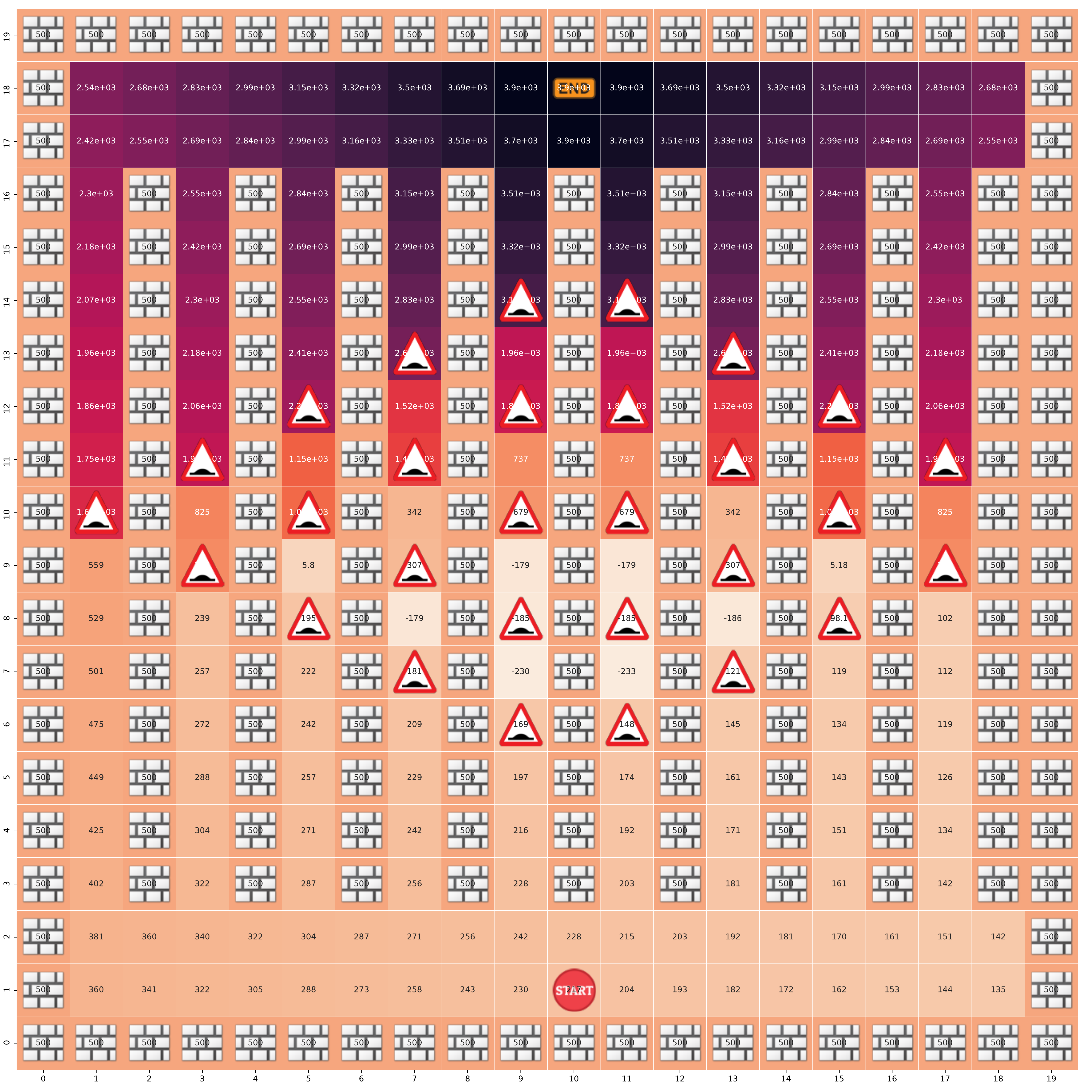}
	% \vspace{-0.7em}
	\caption{State value diagram of a simple maze containing one type of obstacle}
	%\caption{FHE protects against network insecurities in untrusted cloud services, enabling users to securely offload sensitive data}
	\label{fig:simple_maze_state_values}
\vspace{-1.9em}
\end{figure}

Our rewards policy is represented as a cumulative sum of weighted rewards. Auto-tuning rewards policies involve adjusting the parameters of a rewards function based on the observed outcomes of an agent's actions~\cite{fan2018auto, niknia2022nanoscale}. This can be represented as a cumulative sum of weighted rewards, where each individual reward is assigned a weight that reflects its relative importance. The weights are typically adjusted over time based on the agent's performance, with the goal of maximizing the overall cumulative sum of rewards. This approach allows for flexible and adaptive rewards policies that can be tailored to the specific goals and constraints of a given task or environment. Additionally, this method can be used to optimize the trade-off between exploration and exploitation~\cite{sledge2017balancing, liu2021stochastic}, balancing the need to explore new actions with the desire to exploit known successful actions.

Auto-tuning rewards policies have been shown to be effective in a wide range of applications, including robotics, gaming, and autonomous systems~\cite{colbert2021generating, farooq2021efficient}. In general, the performance of these methods is evaluated by comparing the cumulative sum of rewards achieved by the agent using the auto-tuned policy to that of a fixed or manually-tuned policy. Studies have found that auto-tuning rewards policies can lead to significant improvements in performance, particularly in tasks where the optimal rewards function is not known a priori or is difficult to specify manually~\cite{karn2018dynamic}. For example, in robotic tasks, auto-tuning rewards policies have been used to improve the efficiency and robustness of robotic grasping and manipulation~\cite{ashouri2022mlgoperf, ashouri2022work}. In gaming, auto-tuning rewards policies have been used to create more engaging and challenging opponents~\cite{dong2020intelligent, ganesh2022review, prigent2022methodology}. In autonomous systems, these policies have been used to improve the efficiency and safety of self-driving cars~\cite{svetozarevic2022data}.

\section{Experimental Setup}
\label{sec:exp_setup}
We employ a few examples to help understand and evaluate our approach to enhancing design space exploration for policy iteration.
Our experiments lead with a maze-solving example. We start with a simple multi-lane maze that consists of speed bumps, as seen in Figure~\ref{fig:maze_empty}.
% The problem statement can be visualized in Figure~\ref{fig:maze_empty}.
This is followed by a complex maze-solving problem as seen in Figure~\ref{fig:maze_empty_complex}.

Our goal is to highlight the importance of choosing correct penalties and discount factor $\gamma$ that encapsulates every scenario while still converging on to a solution~\cite{baruah_gnnmark_2021}. Choosing the correct penalties and discount factor (gamma) is crucial when solving a multi-objective maze. The penalties should be chosen to accurately reflect the relative importance of different types of obstacles, while the discount factor should be chosen to balance the trade-off between immediate and long-term rewards. When these values are chosen correctly, the algorithm will be able to converge on a solution that considers all relevant scenarios. However, if the penalties or discount factor are not chosen correctly, the algorithm may converge on a suboptimal solution or may not converge at all. Therefore, it is important to carefully consider the problem at hand and choose appropriate penalties and discount factor to ensure that the algorithm converges on an optimal solution.

One way to ensure that a close-to-optimal reward policy and discount factor is considered is to incorporate an auto-tuner using ordinal regression for design space exploration~\cite{nardi2019practical, shivdikar_smash_2021}, similar to polyhedral mapping exhibited in the works~\cite{pfaffe2019efficient, alvarez2005shape, brehm2017polyhedral}. Ordinal regression is a statistical method that can be used to predict an ordinal variable (a variable that can be ranked) based on a set of predictor variables. In this case, the ordinal variable could be the performance of the algorithm and the predictor variables could be the reward policy and discount factor. By using ordinal regression, we can explore the design space of different reward policies and discount factors and identify those that lead to the best performance. The auto-tuner can then use this information to automatically adjust the reward policy and discount factor in real-time to ensure that the algorithm is always operating in a region of the design space that is likely to lead to good performance.

\begin{figure}[htbp]
	\centering
	\includegraphics[width=0.48\textwidth]{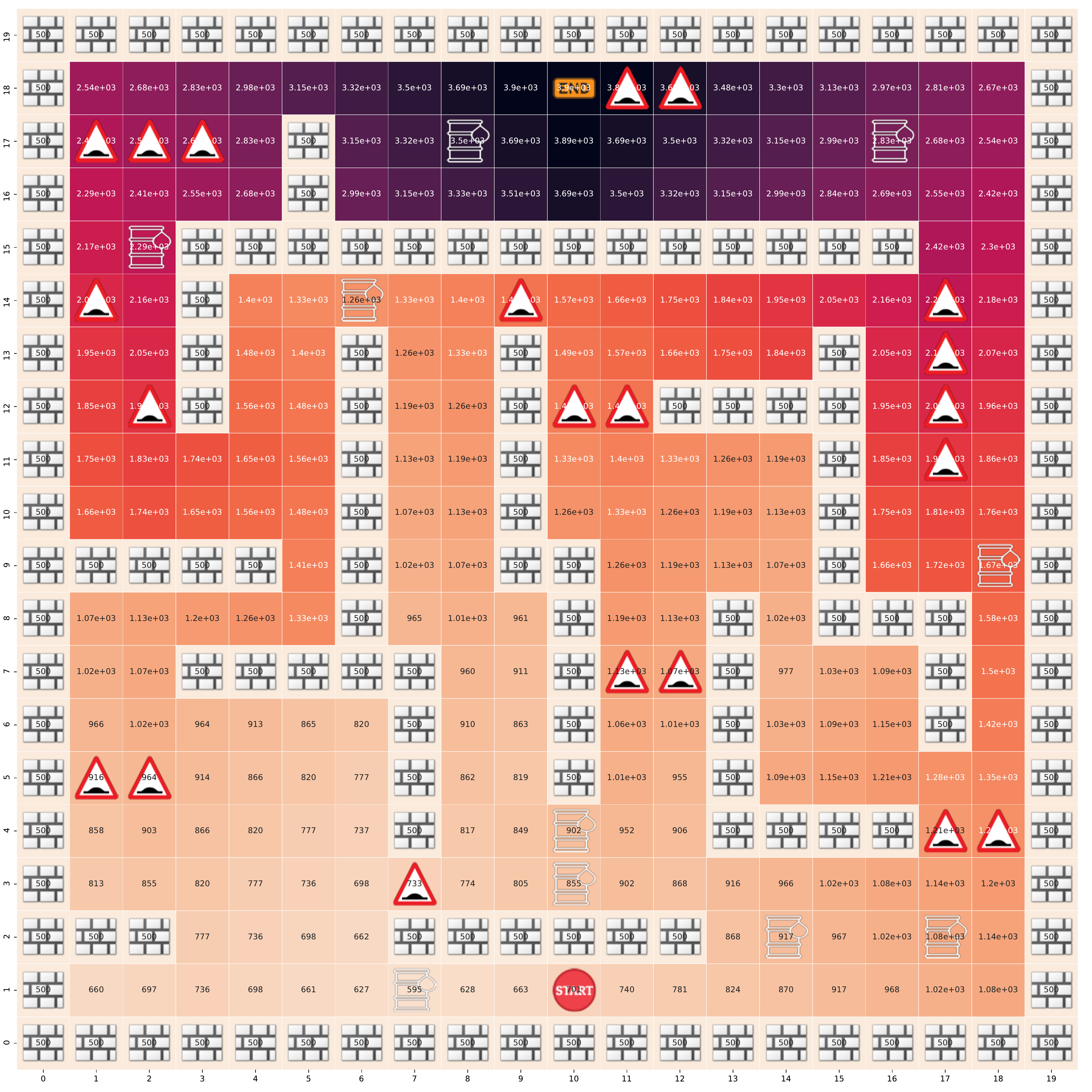}
	% \vspace{-0.7em}
	\caption{State value diagram of a complex maze constituting different types of obstacles}
	%\caption{FHE protects against network insecurities in untrusted cloud services, enabling users to securely offload sensitive data}
	\label{fig:complex_maze_state_values}
\vspace{-1.9em}
\end{figure}

Incorporating our policies, we generate a state-value diagram that represents the value of each state in the maze~\cite{niu2018generalized, niknia2021aging}. The state-value diagram is a graphical representation of the algorithm's policy. Each state in the maze is represented by a box and the edges represent the possible actions that can be taken from that state. The value of each state is represented by a color code, where the darker the color, the higher the value. The state-value diagram is generated by running the algorithm using our proposed policies and comparing the results. It can be used to analyze the performance of the algorithm and identify areas where the algorithm performed well or poorly. This visualization allows for a detailed analysis of the algorithm's behavior and the impact of the policies on the algorithm's performance.
Figures~\ref{fig:simple_maze_state_values} and \ref{fig:complex_maze_state_values} illustrate a heatmap of the state-value diagram for simple and complex mazes, respectively.

\begin{figure}[htbp]
	\centering
	\includegraphics[width=0.48\textwidth]{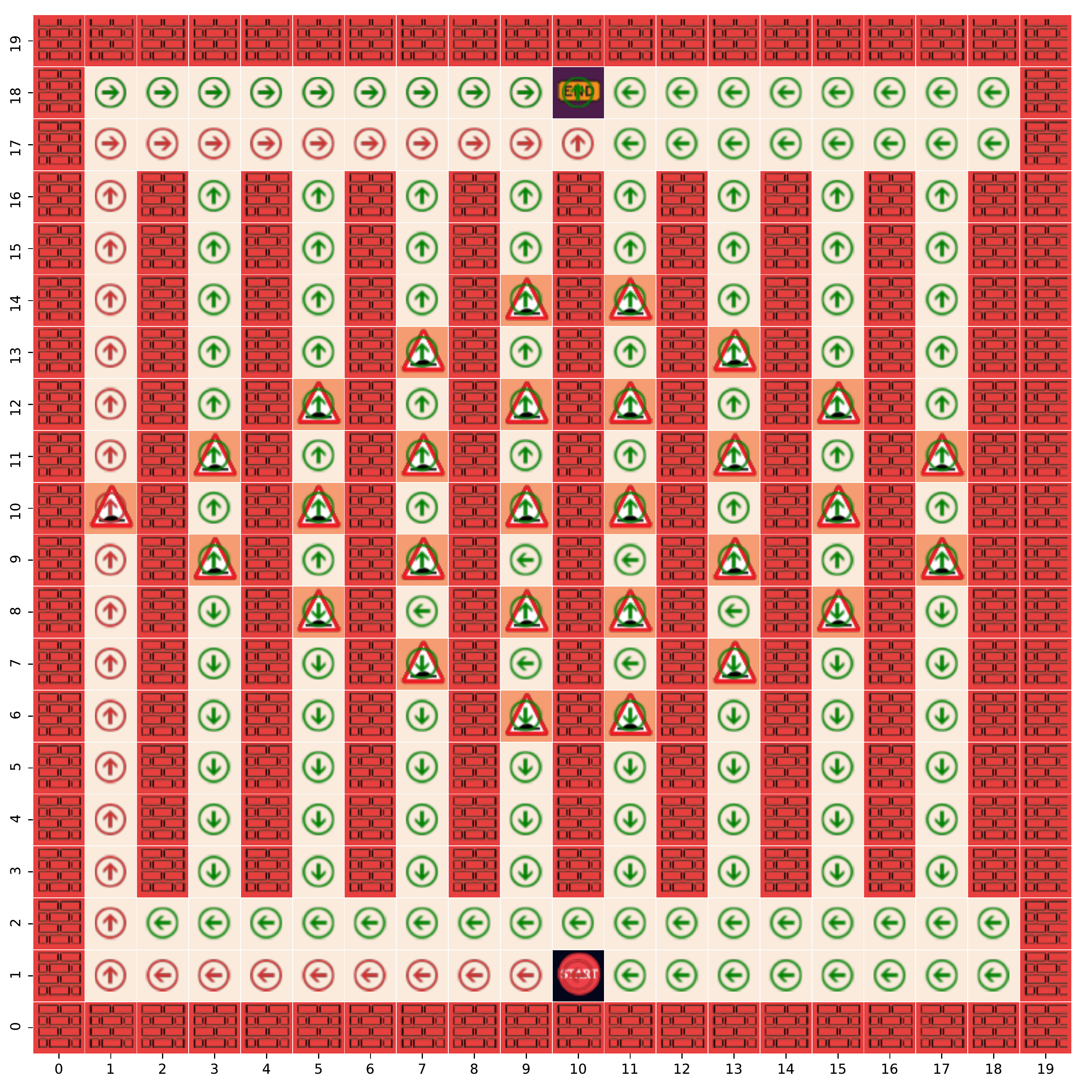}
	% \vspace{-0.7em}
	\caption{High Penalty Simple Maze Path: A visual representation of the algorithm's path using a high penalty policy on a simple maze}
	%\caption{FHE protects against network insecurities in untrusted cloud services, enabling users to securely offload sensitive data}
	\label{fig:high_penaty_simple_path}
% \vspace{-1.9em}
\end{figure}

\begin{figure}[htbp]
	\centering
	\includegraphics[width=0.48\textwidth]{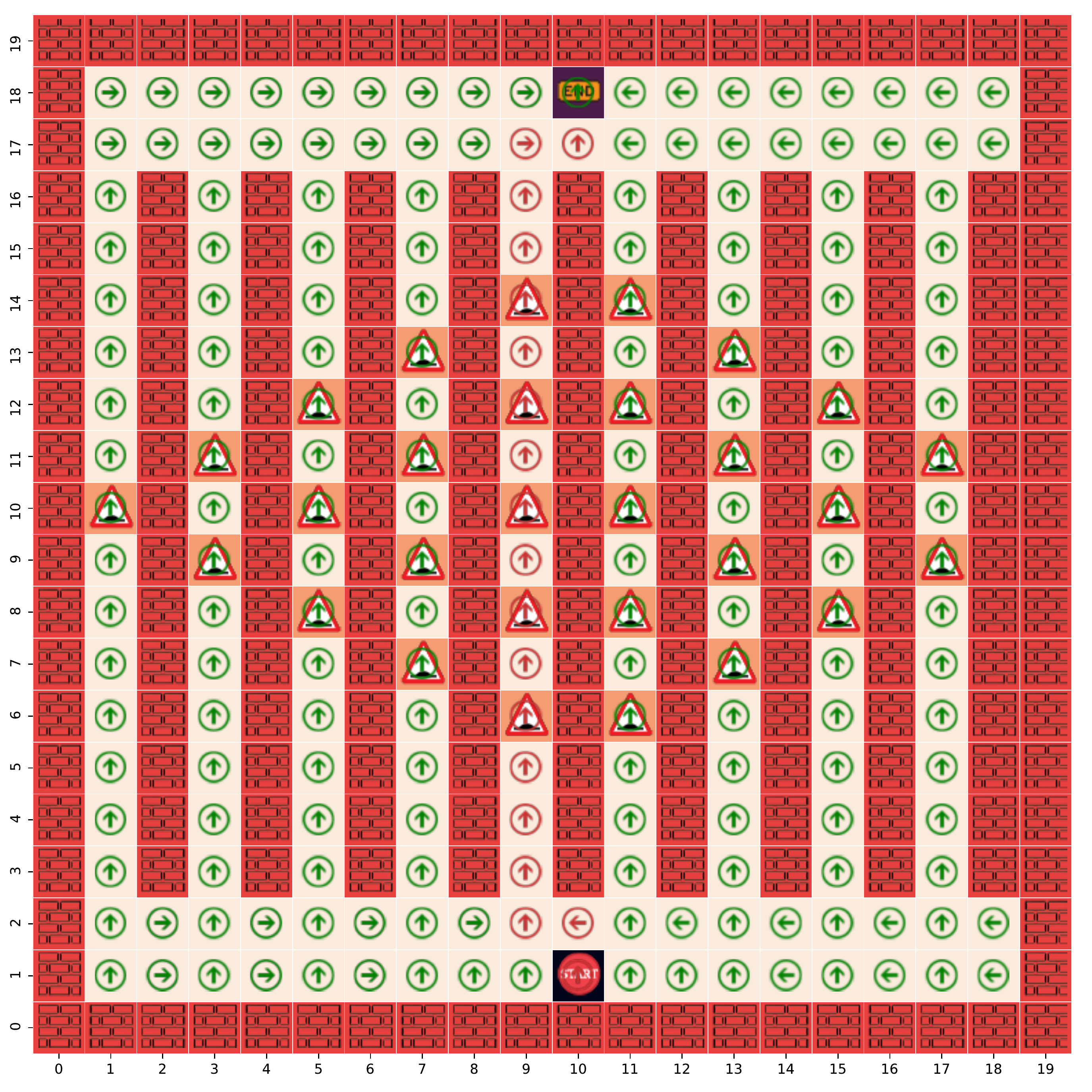}
	% \vspace{-0.7em}
	\caption{Low Penalty Simple Maze Path: Visual representation of the algorithm's path using a low penalty policy}
	%\caption{FHE protects against network insecurities in untrusted cloud services, enabling users to securely offload sensitive data}
	\label{fig:low_penaty_simple_path}
\vspace{-1.0em}
\end{figure}

Our first proposed policy is designed not to penalize obstacles heavily. This means that when the algorithm encounters an obstacle, it will not be heavily punished for doing so. Instead, the obstacle is viewed as a minor inconvenience that does not greatly impact the overall performance of the algorithm~\cite{lee2018gated, srinivasan_dynamic_2016}. This approach is beneficial because it allows the algorithm to explore more of the maze and potentially discover new paths that may lead to a higher reward. However, it also means that the algorithm may take longer to converge on a solution because it is not heavily discouraged from exploring suboptimal paths~\cite{wang_deep_2020, wang_deep_2021}. Additionally, this policy may not work well in situations where certain obstacles are particularly dangerous or costly to encounter. In those cases, a more aggressive penalty may be necessary to ensure that the algorithm avoids those obstacles. This proposed policy can be visualized in Figure~\ref{fig:low_penaty_simple_path} where the maze is shown with different obstacles, and the algorithm's path is traced for the low penalty reward policy.

On the contrary, our second proposed policy heavily penalizes obstacles. This means that when the algorithm encounters an obstacle, it is heavily punished for doing so. This approach is beneficial in situations where certain obstacles are particularly dangerous or costly to encounter. The heavy penalty discourages the algorithm from exploring suboptimal paths and encourages it to find the optimal path more quickly. However, this policy may not work well in situations where the maze is large and complex, and the algorithm may miss potentially useful paths if it avoids obstacles at all costs. Additionally, this policy may lead to suboptimal solutions if the penalties are not chosen correctly. A visual representation of this policy would show a path with fewer 
obstacles encountered (Figure~\ref{fig:high_penaty_simple_path}).

% KTB Redudnat removed
% \begin{figure}[htbp]
% 	\centering
% 	\includegraphics[width=0.48\textwidth]{figures/05_exp/low_penalty_maze_state_values.pdf}
% 	% \vspace{-0.7em}
% 	\caption{3 Maze state values}
% 	%\caption{FHE protects against network insecurities in untrusted cloud services, enabling users to securely offload sensitive data}
% 	\label{fig:maze_state_values}
% % \vspace{-1.9em}
% \end{figure}

\begin{figure}[thbp]
	\centering
	\includegraphics[width=0.48\textwidth]{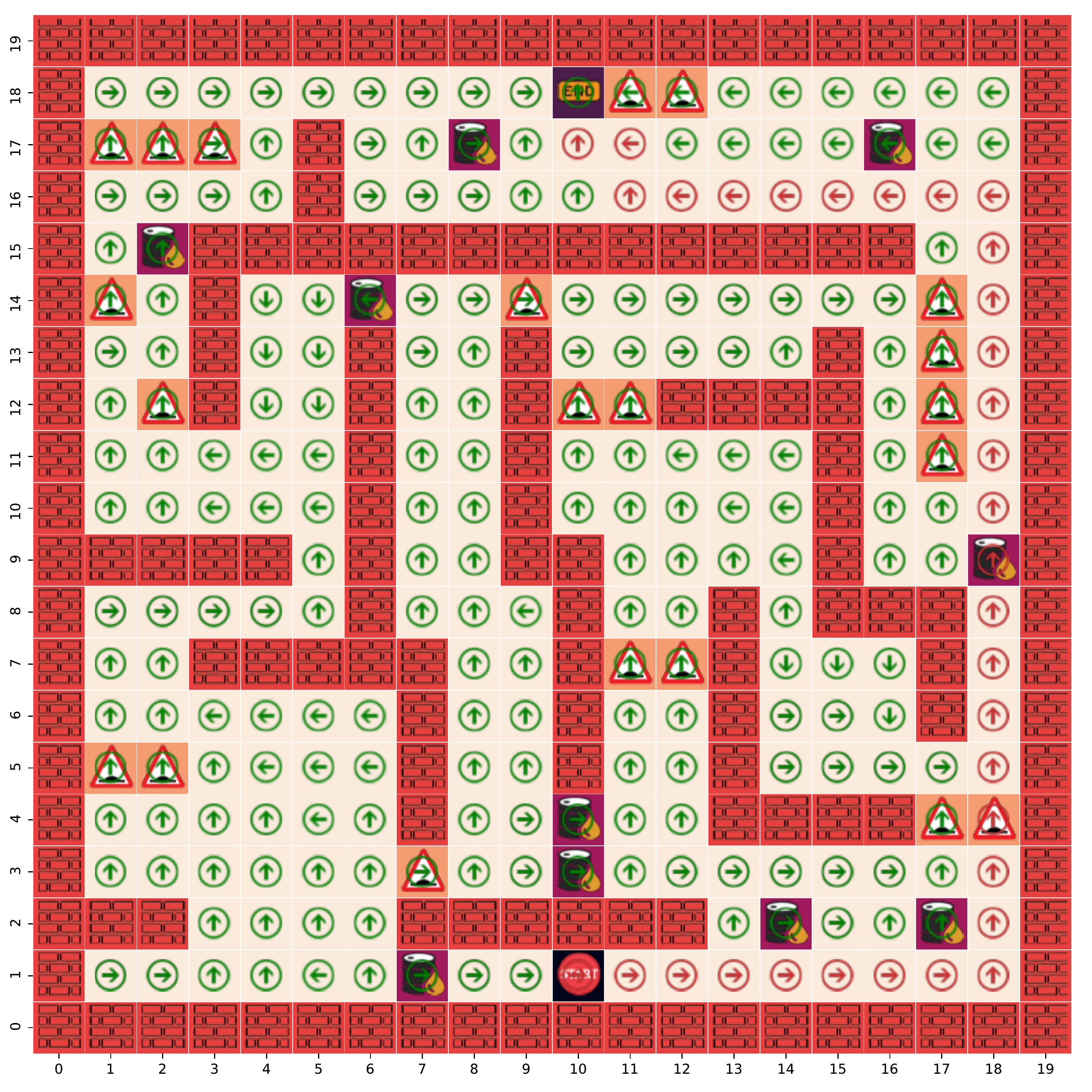}
	% \vspace{-0.7em}
	\caption{Path plot for a multi-modal maze with different types of obstacles}
	%\caption{FHE protects against network insecurities in untrusted cloud services, enabling users to securely offload sensitive data}
	\label{fig:complex_maze_path}
\vspace{-1.9em}
\end{figure}

We perform similar experiments on a complex maze with multiple types of obstacles to see how our proposed policies perform in a more challenging environment. The complex maze includes a variety of obstacles such as walls, speed bumps, and oil spills. Each type of obstacle has a different penalty associated with it. The algorithms are run using both of our proposed policies, and the performance is measured in terms of the final reward obtained, the time taken to reach the final reward, and the number of steps taken to reach the final reward. By comparing the results of the two policies, we can see how well each policy performs in a more complex environment and identify the trade-offs between exploration and exploitation. This experiment allows us to test the robustness of our proposed policies and to evaluate their performance in a more realistic scenario.

Figure~\ref{fig:complex_maze_path} visualizes the best path the algorithms converged on, when using both proposed policies on the complex maze with multiple types of obstacles. The figure shows a detailed view of the maze, including the different types of obstacles and the corresponding penalties. The best path, represented by the path with the highest reward and the least number of steps, is highlighted in red.

% \section{Evaluation}
% \label{sec:eval}
% Evaluation...

% \input{sections/07_performance_analysis}

\section{Results}
\label{sec:results}

% The results of our experiments on the multi-objective mazes show that both of our proposed policies were able to converge on a solution, although with different trade-offs. The first policy, which does not heavily penalize obstacles, was able to explore more of the maze and discover new paths that may lead to a higher reward. However, it also took longer to converge on a solution and the final reward obtained was not as high as the second policy. The second policy, which heavily penalizes obstacles, was able to find the optimal path more quickly, but it also explored less of the maze and missed potentially useful paths. The final reward obtained was higher with this policy. The state-value diagram helped to analyze the algorithm's behavior and the impact of the policies on the algorithm's performance, and to identify areas where the algorithm could be improved. Overall, the results suggest that the choice of the policy and penalties is important and should be chosen based on the specific scenario and the desired trade-offs between exploration and exploitation.

We present the results of our evaluation of eight large mazes of the same size for path planning using policy iteration. Our approach employed an auto-tuner using ordinal regression for design space exploration, with the goal of finding optimal parameters that maximize the performance of the learning agent. We observed that the accumulated rewards varied significantly across the different reward policies and mazes.

\begin{figure*}[htbp]
	\centering
\includegraphics[width=1.0\textwidth]{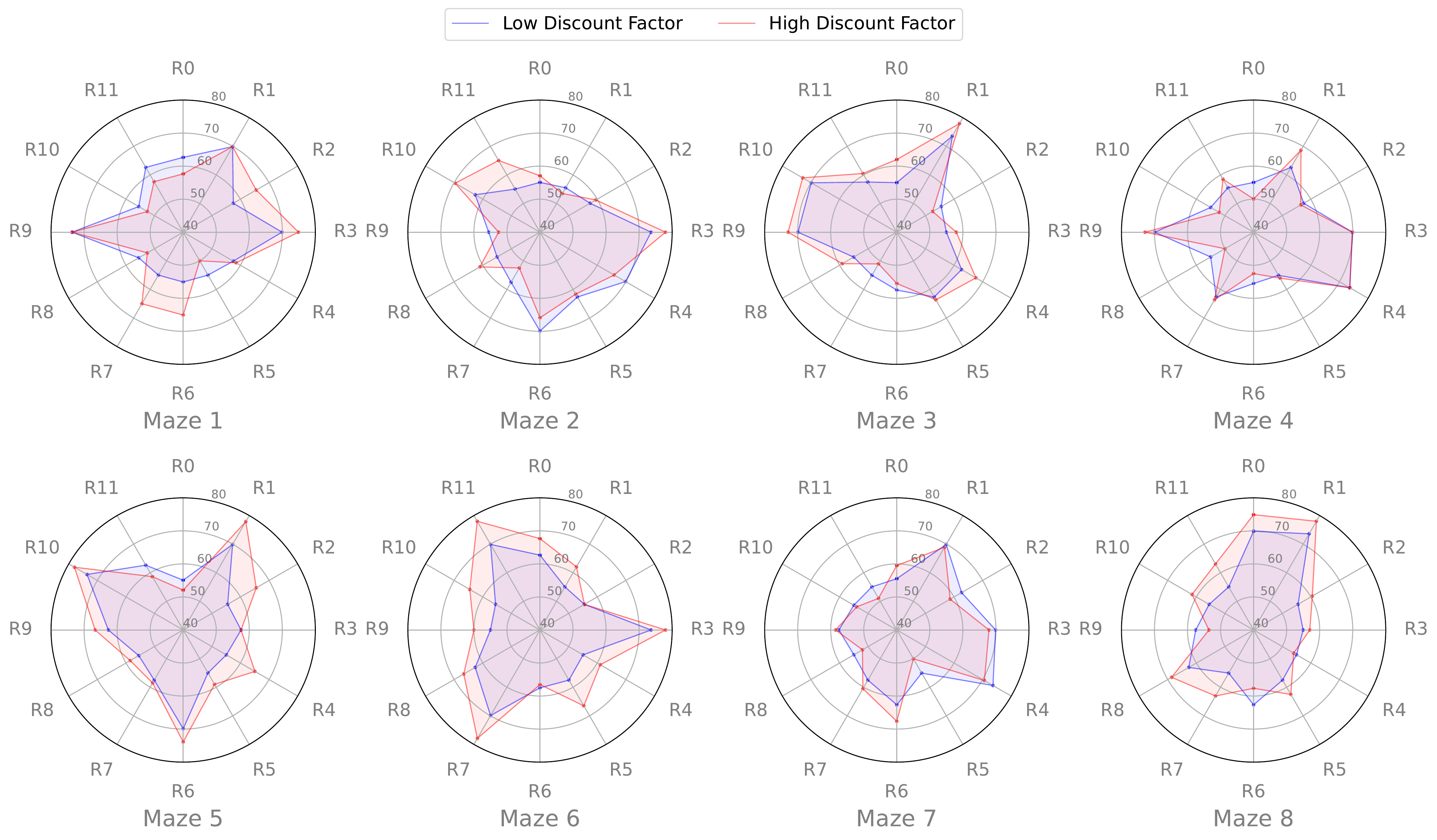}
	\vspace{-1.6em}
	\caption{Comparing Accumulated Rewards and Discount Factors Across Multiple Mazes using Spider Plots. Each spider plot displays the Top-12 reward policies generated by the Auto-Tuner.}
	\label{fig:maze_radar}
\vspace{-1.6em}
\end{figure*}

We plotted the accumulated rewards for each maze for twelve reward policies ($R0 - R11$) generated by the auto-tuner. This has been visualized as a spider plot in Fig~\ref{fig:maze_radar}. The results indicated that the reward policy that accumulated the highest reward differed from maze to maze. This suggests that the optimal reward policy is heavily dependent on the environment in which the learning agent is operating. In addition, certain mazes consistently generated higher accumulated rewards across all policies when the discount factor was low. This indicates that far-sighted policies outweigh the gains of the near-sighted policies for these mazes.

We also experimented with high and low discount factors to understand the effects of near-sighted v/s far-sighted policy iteration. Our results indicated that the choice of discount factor had a significant impact on the learning agent's performance. We observed that the performance of the learning agent was better when the discount factor was low for certain mazes, while for others, the performance was better when the discount factor was high.

In summary, our evaluation demonstrated the importance of carefully selecting the reward policy and discount factor in policy iteration for path planning in complex environments. The results indicate that the optimal parameters for the learning agent depend on the specific characteristics of the environment in which it operates. Future work will explore additional approaches for optimizing the learning agent's performance in complex environments, including the use of more advanced machine learning algorithms and more complex mazes.

Our approach can be applied to a wide range of path-planning problems in real-world applications, including autonomous vehicles, robotics, and logistics. Our results provide valuable insights for designing efficient and effective learning agents for path planning in complex environments.

% The spider plot of accumulated rewards for 12 different reward policies (R0 to R11) is a visual representation of the performance of these policies in the context of policy iteration parameter fine-tuning using an auto-tuner. The plot compares the accumulated rewards of each policy under two different discount factors - low and high. The spider plot is a useful tool for quickly assessing the performance of each policy and identifying which ones perform better under certain conditions.

% The spider plot shows that some reward policies perform better than others under low and high discount factors. For example, policies R3, R4, and R11 perform well under both low and high discount factors, while policies R0 and R10 perform poorly under both conditions. This suggests that certain reward policies are more robust to changes in discount factor and may be better suited for different applications where the discount factor may vary.

In the context of policy iteration parameter fine-tuning using an auto-tuner, Fig~\ref{fig:maze_radar} can be used to guide the exploration of the reward function design space. By identifying the policies that perform well under different discount factors, the auto-tuner can focus on exploring reward functions that are similar to those policies. This can help speed up the tuning process and improve the overall performance of the policy.

Overall, the spider plot of accumulated rewards is a valuable tool for evaluating the performance of different reward policies in the context of policy iteration parameter fine-tuning using an auto-tuner. By identifying which policies perform well under different conditions, the plot can guide the exploration of the reward function design space and help to improve the performance of the policy.

\section{Related Work}
\label{sec:related_work}
% Related Work...

The problem of path planning in complex environments has been extensively studied in the field of robotics and artificial intelligence~\cite{sariff2006overview, yao2020path, wu2019tdpp}. In recent years, there has been a growing interest in using reinforcement learning (RL) approaches for path planning, which allows agents to learn from their own experiences and improve their performance over time. RL-based path planning methods have been applied to a wide range of real-world applications, including autonomous vehicles, mobile robots, and unmanned aerial vehicles.

A popular RL approach for path planning is policy iteration, which involves iteratively evaluating and improving a policy for selecting actions in a given state. In policy iteration, the choice of a reward function and discount factor can significantly impact the performance of the learning agent. Several studies have explored different reward functions and discount factors for path planning in complex environments, with the goal of improving the performance of the learning agent.

The use of auto-tuners for design space exploration has been gaining popularity in recent years, particularly in the field of machine learning. Auto-tuners use optimization algorithms to search the design space for optimal hyperparameters that maximize the performance of the learning algorithm. Several studies have explored the use of auto-tuners for optimizing the performance of RL-based path-planning algorithms. These studies have demonstrated that auto-tuners can significantly improve the performance of RL-based path-planning algorithms, particularly in complex environments with a large number of hyperparameters.

In contrast to our study, which focuses on the selection of optimal hyperparameters for policy iteration in path planning, the work of Li et al. in~\cite{li2009online} addresses the problem of balancing exploration and exploitation in large or continuous Markov decision processes. The work proposes a practical solution to exploring large MDPs by integrating a powerful exploration technique, Rmax, into a state-of-the-art learning algorithm, least-squares policy iteration (LSPI). This approach combines the strengths of both methods and has shown its effectiveness and superiority over LSPI with two other popular exploration rules in several benchmark problems. While both studies use policy iteration for path planning in complex environments, our study specifically focuses on the selection of optimal hyperparameters for the learning agent, while the work~\cite{li2009online} addresses the problem of exploration in large or continuous MDPs.

This work focuses on the selection of optimal hyperparameters for policy iteration in path planning, Zamfirache et al., in their work~\cite{zamfirache2022policy}, propose a new reinforcement learning-based control approach that uses the policy iteration and a metaheuristic Grey Wolf Optimizer (GWO) algorithm to train neural networks for solving complex optimization problems. The paper presents experimental results comparing the proposed approach to the classical policy iteration reinforcement learning-based control approach using the Gradient Descent (GD) algorithm and an approach that uses the Particle Swarm Optimization (PSO) algorithm~\cite{thakkar_video_2017}. The experiments were conducted using a nonlinear servo system laboratory equipment to evaluate how well each approach solves the optimal reference tracking control for an experimental servo system position control system. The study shows that the GWO algorithm represents a better solution compared to GD and PSO algorithms for the control objective considered in the paper. While both studies use policy iteration for reinforcement learning, the focus of the research~\cite{zamfirache2022policy} is on using metaheuristic optimization algorithms to improve the performance of the learning agent, whereas our study focuses on optimizing the hyperparameters for policy iteration in path planning.

Gao et al. in their work~\cite{gao2021reinforcement} use policy iteration for reinforcement learning; the focus of the this paper is on addressing the challenges in developing efficient and effective learning control algorithms for human-robot systems. The study introduces the flexible policy iteration (FPI) algorithm, which can flexibly and organically integrate experience replay and supplemental values from prior experience into the reinforcement learning controller. The study provides system-level performances, including convergence of the approximate value function, optimality of the solution, and stability of the system. The effectiveness of the FPI is demonstrated via realistic simulations of the human-robot system. The paper addresses the challenges of obtaining a customized mathematical model of a human-robot system either online or offline, which makes design methods based on classical control theory nearly impossible. Our approach converges faster to an optimal policy tailored to our problem statement requiring computationally far fewer resources.

\section{Conclusion}
\label{sec:conclusion}

Our evaluation using policy iteration for path planning in complex environments demonstrated the importance of carefully selecting the reward policy and discount factor for optimizing the performance of the learning agent. We employed an auto-tuner using ordinal regression for design space exploration and evaluated eight large mazes of the same size, each with a varying number of obstacles of different types. We plotted the accumulated rewards for each maze for twelve reward policies generated by the auto-tuner and also experimented with high and low discount factors to understand the effects of near-sighted vs far-sighted policy iteration.

This research makes several contributions toward improving Policy Iteration. First, it provides a comprehensive analysis of previous attempts to optimize search space exploration. Second, it evaluates different permutations of learning parameters for the Policy Iteration algorithm. Third, it proposes a novel approach based on ordinal regression auto-tuning to speed up the design-space exploration of reward parameters. Finally, the proposed solution is evaluated on various problem statements, and the results show an average acceleration of 1.48× over heuristic approaches. These contributions collectively advance the field of Policy Iteration and can help accelerate the optimization process for a wide range of applications.

Our results indicate that the optimal parameters for the learning agent depend on the specific characteristics of the environment in which it operates. The spider plots of accumulated rewards for the top 12 reward policies, comparing both low and high discount factors for eight different mazes of the same size provide valuable insights for designing efficient and effective learning agents for path planning in complex environments. Future work will explore additional approaches for optimizing the learning agent's performance in complex environments, including the use of more advanced machine learning algorithms and more complex mazes. Our approach can be applied to a wide range of path-planning problems in real-world applications, including autonomous vehicles, robotics, and logistics.

\section*{Acknowledgements}

We would like to express our gratitude to all the individuals and organizations who have contributed to the completion of this study. We would like to thank the members of our research team for their hard work and dedication to this project.

We would like to thank our funding agency for providing the financial support necessary to carry out this research. Additionally, we would like to acknowledge the developers of the software and tools used in this study, as their contributions were essential to our work. Finally, We would like to extend our gratitude to the reviewers who provided valuable feedback and constructive criticism that helped us to improve the quality of this study.
%We hope that our findings will contribute to the advancement of the field of reinforcement learning-based path planning, and we look forward to continuing our research in this area.
\balance
\bibliographystyle{IEEEtran}
%\balance
\bibliography{bibliography/biblio, bibliography/zotero_ref}

\end{document}